\documentclass[lettersize,journal]{IEEEtran}
\usepackage{amsmath,amsfonts}
\usepackage{algorithm}
\usepackage{algorithmic}
\usepackage{array}
\usepackage[caption=false,font=normalsize,labelfont=sf,textfont=sf]{subfig}
\usepackage{textcomp}
\usepackage{stfloats}
\usepackage{url}
\usepackage{verbatim}
\usepackage{graphicx}
\usepackage{multirow}
\usepackage{pifont}
\usepackage{adjustbox}
\usepackage{booktabs}
\usepackage[table,xcdraw]{xcolor}
\def\BibTeX{{\rm B\kern-.05em{\sc i\kern-.025em b}\kern-.08em
    T\kern-.1667em\lower.7ex\hbox{E}\kern-.125emX}}
\usepackage{balance}
\begin{document}

\title{High-Fidelity One-Step Generative Visuomotor Policy via Recursive Correction, Frequency Consistency, and Contrastive Flow Matching}

\author{Yuran Chen, Xinye Cai, Zhonglin Gong, and Yang Huang%
\thanks{This work was supported in part by the National Natural Science Foundation of China under Grants 62572005 and U24A20267. \textit{(Corresponding author: Xinye Cai.)}}%
\thanks{Yuran Chen, Zhonglin Gong, and Yang Huang are with the School of Safety Science and Engineering, Anhui University of Science and Technology, Huainan 232001, China, and also with the State Key Laboratory of Digital Intelligent Technology for Unmanned Coal Mining, Anhui University of Science and Technology, Huainan 232001, China (e-mail: 1014913959@qq.com).}%
\thanks{Xinye Cai is with the State Key Laboratory of Digital Intelligent Technology for Unmanned Coal Mining, Anhui University of Science and Technology, Huainan 232001, China, and also with the School of Artificial Intelligence, Anhui University of Science and Technology, Hefei 231131, China (e-mail: xinye@aust.edu.cn).}%
}

\markboth{Preprint submitted to arXiv}
{Chen \MakeLowercase{\textit{et al.}}: High-Fidelity One-Step Generative Visuomotor Policy}

\maketitle
\begin{abstract}
Generative models such as Diffusion and Flow Matching have significantly advanced robotic visuomotor policies by modeling complex and multimodal action distributions. However, their reliance on multi-step sampling or ODE solvers introduces substantial inference latency. Existing single-step acceleration methods improve efficiency, but they often approximate the entire iterative action-generation process with a single large update, introducing an approximation gap. We characterize this gap as three coupled failure modes: spatial deviation, where large-step approximation ignores local flow geometry; frequency distortion, where over-smoothed predictions suppress fine-grained manipulation details; and mode averaging, where entangled multimodal flows produce ambiguous actions. To address these challenges, we propose a high-fidelity one-step generative visuomotor policy framework based on three complementary mechanisms: recursive correction, frequency consistency, and contrastive flow matching. First, \textbf{Recursive Consistent Action Flow (RCAF)} performs high-order recursive correction to compensate for spatial truncation errors, aligning one-step predictions with recursively refined flow trajectories. Second, \textbf{Dual-Timestep Frequency Consistency (DTFC)} preserves high-frequency manipulation details by encouraging adaptive spectral consistency across different flow timesteps. Third, \textbf{Contrastive Flow Matching (CFM)} separates entangled action flows with a margin-based repulsive objective, reducing mode averaging and action ambiguity in multimodal manipulation scenarios. Extensive experiments on RoboTwin, RoboTwin 2.0, Adroit, DexArt, and real-world robot platforms show that our method achieves competitive or superior manipulation performance compared with strong 10-step generative policy baselines while requiring only a single forward pass (1 NFE), reducing inference cost for low-latency visuomotor control.
\end{abstract}


\section{Introduction}
Visual imitation learning~\cite{levine2016end,zheng2022imitation} enables robots to acquire 
manipulation skills by learning direct mappings from observations to 
actions from expert demonstrations. As an important paradigm in robot learning, it has shown promising results in a wide range of manipulation scenarios, including dexterous grasping, precise assembly, bimanual coordination, and 
long-horizon object manipulation~\cite{chi2025diffusion,rajeswaran2017learning,
ze20243d,liu2024rdt,black2024pi_0,chen2025robotwin,liu2024visuomotor,wang2022learning}. A central challenge in 
this setting is that expert action distributions are often high-dimensional, 
temporally correlated, and multimodal, since multiple valid action sequences may solve the same task under the same visual observation.

Recently, generative models based on continuous-time dynamics, such as diffusion models~\cite{ho2020denoising,song2020denoising} and flow matching 
models~\cite{lipman2022flow,liu2022flow}, have become effective tools for 
visuomotor policy learning. By modeling action generation as a gradual 
transformation from noise to expert actions, these methods can capture 
multimodal action distributions and generate temporally coherent action 
trajectories conditioned on visual observations~\cite{chi2025diffusion,ze20243d,
chisari2024learning,liu2024rdt}. However, their high generation quality usually 
comes with substantial inference latency. During deployment, standard diffusion 
or flow-based policies require multiple denoising steps or ODE solver 
evaluations to produce one action chunk. This iterative inference process limits their applicability to high-frequency closed-loop control, where the robot must react quickly to changing observations and contact dynamics.

To reduce inference cost, recent studies have explored single-step or few-step 
action generation strategies, including knowledge distillation 
methods~\cite{lu2024manicm,wang2024one}, mean-flow-based approaches 
~\cite{sheng2025mp1,zou2025dm1,zhan2026mean}, and consistency-based policies 
~\cite{yan2025maniflow,prasad2024consistency,fang2025imitation,
zhang2025flowpolicy}. While these methods improve sampling efficiency, they often compress a complex iterative generation process into a single large update. We argue that the main difficulty of single-step visuomotor generation lies in the approximation gap introduced by this compression. In high-dimensional manipulation, expert action distributions are not only multimodal, but also lie on nonlinear manifolds with fine-grained temporal and spatial structures. Therefore, a single large update must simultaneously approximate local flow geometry, preserve high-frequency motion details, and distinguish different valid action modes, which is challenging for precise contact-rich control.

\begin{figure*}[!t]
    \centering
    \includegraphics[width=\textwidth,
    clip]{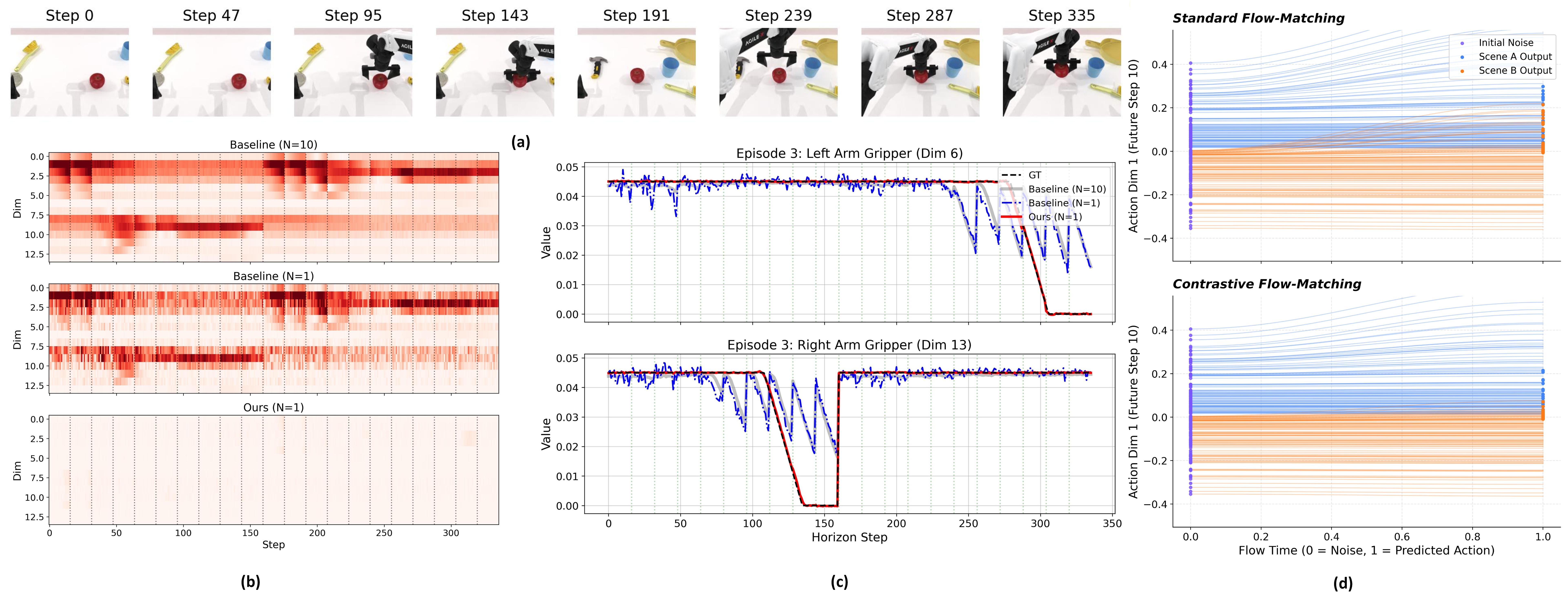}
    \vspace{-8mm}
    \caption{Qualitative analysis of the single-step generation models. (a) Task sequence: Visual progression of the bimanual manipulation task. (b) Joint error heatmaps: Compared to baselines, our method (Ours, N=1) exhibits lower joint angle errors, helping mitigate spatial truncation errors. (c) High-frequency gripper tracking: Plagued by error accumulation, the baseline (blue line) suffers from larger high-frequency oscillations and instability. Conversely, our model (red line) better preserves transient dynamics with improved high-frequency stability. (d) Flow space trajectories: Contrastive Flow Matching (bottom) helps decouple the intertwined vector fields seen in the standard method (top) into independent paths, mitigating mode averaging.}
    \label{fig:1}
\end{figure*}

As illustrated in Figure~\ref{fig:1}, this approximation gap manifests as three 
common failure modes. First, a large integration step can ignore local manifold 
geometry, causing spatial truncation errors (Figure~\ref{fig:1}(b)). Second, 
optimizing a single long-span prediction can encourage over-smoothed outputs, 
suppressing high-frequency manipulation details that are important for precise 
control (Figure~\ref{fig:1}(c)). Third, when multiple valid action modes are 
entangled in the same flow space, the model may produce averaged or ambiguous 
actions, reducing the reliability of generated trajectories in multimodal 
manipulation scenarios (Figure~\ref{fig:1}(d)).

To address these limitations, we propose a high-fidelity one-step generative visuomotor policy framework that targets the approximation gap between iterative action generation and one-step inference. Consistent with this objective, the framework integrates three complementary mechanisms: recursive correction for compensating spatial truncation errors, frequency consistency for preserving fine-grained manipulation details, and contrastive flow matching for separating multimodal action flows.

First, \textbf{Recursive Consistent Action Flow (RCAF)} performs high-order recursive correction to compensate for spatial truncation errors induced by large-step approximation. Second, \textbf{Dual-Timestep Frequency Consistency (DTFC)} enforces adaptive spectral consistency across different flow timesteps to reduce frequency distortion and preserve high-frequency manipulation details. Third, \textbf{Contrastive Flow Matching (CFM)} introduces a margin-based repulsive objective to separate conflicting action flows, thereby alleviating multimodal flow entanglement and mode averaging. Together, these designs improve the fidelity of one-step action generation while maintaining the efficiency of single-step inference.

We evaluate the proposed framework on a diverse set of simulation benchmarks and real-world robotic platforms. The simulation experiments cover bimanual 
manipulation, cross-domain generalization, and dexterous control tasks from 
RoboTwin~\cite{mu2025robotwin}, RoboTwin 2.0~\cite{chen2025robotwin}, 
Adroit~\cite{rajeswaran2017learning}, and DexArt~\cite{bao2023dexart}. 
Furthermore, we deploy our method on real robotic platforms, including a 
low-cost SO101 platform and an industrial UR7E collaborative robot. 
Experimental results show that our method achieves competitive or better 
manipulation success rates than strong multi-step generative policy baselines 
while using only a single function evaluation. 

The main contributions of this paper are summarized as follows:
\begin{itemize}
\item We propose \textbf{Recursive Consistent Action Flow (RCAF)}, a high-order recursive correction mechanism that compensates for spatial deviations induced by long-step approximation.
\item We introduce \textbf{Dual-Timestep Frequency Consistency (DTFC)} and \textbf{Contrastive Flow Matching (CFM)} to preserve high-frequency manipulation details and separate multimodal action flows, respectively.
\item We validate the proposed framework on diverse simulation benchmarks and real-world robotic platforms, demonstrating competitive manipulation performance with only one function evaluation.
\end{itemize}

The remainder of this paper is organized as follows: Section 2 reviews related work in visuomotor policy and generative policies. Section 3 introduces the problem formulation and necessary preliminaries. Section 4 details the proposed recursive correction, frequency consistency, and contrastive flow matching mechanisms, along with the overall training objectives. Section 5 presents the quantitative and qualitative experimental analyses in both simulation environments and on real-world robots. Finally, Section 6 concludes the paper and discusses future research directions.

\section{Related Work}
\label{sec:related_work}

\subsection{Generative Visuomotor Policies}

Imitation learning aims to acquire control policies directly from expert demonstrations. Traditional behavioral cloning methods often struggle with multimodal expert data, where averaging over multiple valid actions can lead to ambiguous or suboptimal predictions. Recently, generative models have become increasingly important for visuomotor policy learning due to their ability to represent complex and multimodal action distributions. Representative methods such as Diffusion Policy~\cite{chi2025diffusion} 
and its 3D extensions~\cite{ze20243d,ke20243d} formulate action generation as a conditional iterative denoising process, often combined with action chunking for temporally coherent control. Flow Matching (FM)~\cite{lipman2022flow,liu2022flow} provides a continuous-time alternative by learning deterministic ODE paths between noise and expert actions. Compared with diffusion-based sampling, FM can produce smoother transport paths and has been adopted in various dexterous manipulation settings~\cite{chisari2024learning,gkanatsios20253d}. In parallel, large-scale Vision-Language-Action (VLA) models, such as RDT~\cite{liu2024rdt} and $\pi_0$~\cite{black2024pi_0}, further combine generative action modeling with large-scale pretraining to improve cross-task generalization.

Despite their strong modeling capacity, these generative visuomotor policies 
typically rely on iterative denoising or ODE integration during inference. This multi-step generation process introduces non-negligible latency, making it challenging to deploy such policies in high-frequency closed-loop robotic control, where actions must be produced quickly in response to changing observations and contact dynamics.

\subsection{Accelerated Visuomotor Policies}

To reduce the inference cost of iterative generation, recent studies have explored several acceleration paradigms. Knowledge distillation and consistency-based methods~\cite{song2023consistency,lu2024simplifying,sunany} learn to approximate multi-step generative processes with fewer sampling steps, achieving significant speedups in robotic manipulation~\cite{prasad2024consistency,lu2024manicm,
wang2024one}. These ideas have also been integrated into Flow Matching 
frameworks~\cite{yan2025maniflow,yang2024consistency}, with recent methods 
introducing multi-step consistency constraints to improve training stability 
~\cite{fang2025imitation}. However, directly compressing a curved iterative 
generation process into a one-step approximation may introduce spatial deviation when local action-flow geometry is not well preserved.

Another line of work studies mean-flow-based acceleration. Mean Flow 
methods~\cite{geng2025mean,geng2025improved} and their robotic adaptations 
~\cite{sheng2025mp1,zou2025dm1,zhan2026mean} directly learn interval-averaged 
velocities, enabling efficient single-NFE generation. While effective for reducing sampling cost, the interval-average formulation may be limited when approximating highly nonlinear action manifolds over long temporal spans. Warm-start approaches~\cite{wang2024sparse,jiang2025streaming,li2026step} instead use historical trajectories as priors to shorten the sampling path, but they still require iterative refinement at inference time. Other auxiliary techniques, such as Conditional Optimal Transport~\cite{sochopoulos2025fast} and frequency-aware alignment~\cite{su2025freqpolicy}, attempt to improve the geometry or temporal structure of the generative process.

Overall, existing acceleration methods mainly focus on reducing the number of 
sampling steps, while the approximation gap introduced by single-step inference 
remains insufficiently addressed. In particular, single-step visuomotor policies still face spatial deviation, frequency distortion, and multimodal flow entanglement. Our framework addresses these issues through RCAF, DTFC, and CFM, which respectively improve spatial correction, high-frequency detail preservation, and multimodal action-flow separation for efficient robotic automation.

\section{Preliminaries}

\subsection{Visuomotor Policy}

In visuomotor policy learning, the objective is to learn a conditional policy that maps visual observations and robot states to continuous actions. Given an expert demonstration dataset $\mathcal{D}=\{\tau_i\}_{i=1}^{M}$, each trajectory is represented as $\tau=\{(o_k,a_k)\}_{k=1}^{T}$, where $o_k$ and $a_k$ denote the observation and expert action at control timestep $k$. The observation is encoded as a conditioning feature $c_k$.

Following action-chunking policies, the model predicts a future action sequence rather than a single action. We denote the expert action chunk under condition $c_k$ as $x_1\in\mathbb{R}^{D}$, where $D$ is the flattened action dimension. For simplicity, we omit the control index and write the condition as $c$ hereafter.

\subsection{Flow Matching}

Flow Matching learns a continuous vector field that transports samples from a simple prior distribution to the expert action distribution. Let $x_0\sim\mathcal{N}(0,I)$ denote a noise action sample, and let $x_1\sim q(x_1|c)$ denote an expert action chunk. The generation process is defined by the ODE:
\begin{equation}
\label{eq:ode}
\frac{d x_t}{d t}=v_{\theta}(x_t,t,c), \quad t\in[0,1],
\end{equation}
where $v_{\theta}$ is the learned conditional vector field. A commonly used conditional OT path linearly interpolates between $x_0$ and $x_1$:
\begin{equation}
\label{eq:ot_path}
x_t=(1-t)x_0+t x_1 .
\end{equation}
The corresponding target velocity is $x_1-x_0$, yielding the standard Flow Matching objective:
\begin{equation}
\label{eq:fm_loss}
\mathcal{L}_{\text{FM}}
=
\mathbb{E}_{t\sim\mathcal{U}[0,1],x_0,x_1,c}
\left[
\left\|v_{\theta}(x_t,t,c)-(x_1-x_0)\right\|_2^2
\right].
\end{equation}

\subsection{Consistency Training}

Consistency training learns a function $f_{\theta}(x_t,t,c)$ that maps different states along the same generation trajectory to a consistent endpoint. An exponential moving average (EMA) teacher is commonly used for self-distillation between nearby timesteps:
\begin{equation}
\label{eq:ct_loss}
\mathcal{L}_{\text{CT}}
=
\mathbb{E}_{x_t,t,\Delta t,c}
\left[
\left\|
f_{\theta}(x_t,t,c)
-
\operatorname{sg}\left(
f_{\theta_{\text{EMA}}}(x_{t+\Delta t},t+\Delta t,c)
\right)
\right\|_2^2
\right],
\end{equation}
where $\operatorname{sg}(\cdot)$ denotes the stop-gradient operation.

This formulation enables fast one-step inference by directly mapping an intermediate noisy action state to the final action chunk. However, replacing iterative generation with a one-step approximation may under-model intermediate trajectory corrections, especially over long integration spans. This can introduce spatial deviation from the iterative generation trajectory, motivating our RCAF design for high-order recursive correction.

\section{Methods}

\begin{figure*}[!t]
    \centering
    \includegraphics[width=\textwidth, clip]{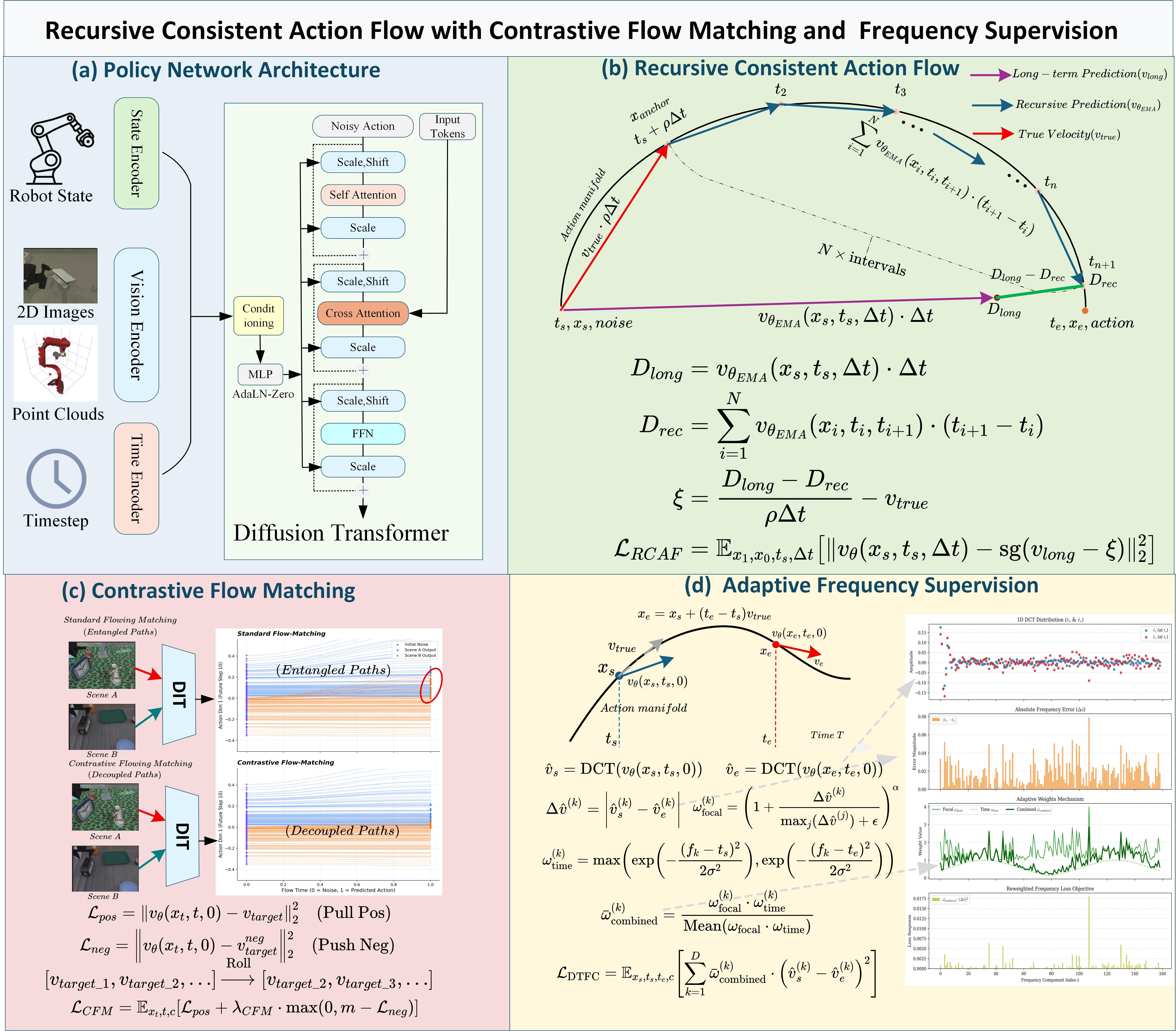}
    \vspace{-0.8cm}
    \caption{Overview of the proposed framework. (a) The multimodal Diffusion Transformer (DiT) policy network architecture. (b) Recursive Consistent Action Flow (RCAF) for mitigating spatial truncation errors via high-order residual correction. (c) Contrastive Flow Matching (CFM) for decoupling interwoven flow fields and resolving mode averaging. (d) Dual-Timestep Frequency Consistency (DTFC) for adaptive high-frequency detail recovery.}
    \label{fig:2}
\end{figure*}

To achieve fast, high-fidelity single-step inference for robotic visuomotor control, we propose a generative policy framework based on recursive correction, frequency consistency, and contrastive flow matching, as illustrated in Figure~\ref{fig:2}. The framework targets three key issues in existing accelerated generative policies: spatial truncation errors, high-frequency detail loss, and multimodal action-flow entanglement.

Specifically, the framework processes multimodal inputs (e.g., proprioception, 2D images, or point clouds) through a unified Diffusion Transformer (DiT) backbone~\cite{peebles2023scalable,yan2025maniflow} equipped with adaptive layer normalization (adaLN) for efficient condition fusion (Figure~\ref{fig:2}a), ultimately predicting a continuous action velocity field. Building upon this generative backbone, we integrate three complementary mechanisms:

First, we introduce Recursive Consistent Action Flow (RCAF) (Section~\ref{sec:rcaf}), which performs recursive correction by distilling multi-step manifold curvature into a single-step residual direction (Figure~\ref{fig:2}b). 
Second, to prevent the policy from degenerating into low-frequency, overly smoothed motions, we propose Dual-Timestep Frequency Consistency (DTFC) (Section~\ref{sec:dtfc}), which preserves important high-frequency manipulation details through adaptive spectral supervision (Figure~\ref{fig:2}d). 
Finally, to reduce the action ambiguity caused by multimodal distribution overlap, we formulate Contrastive Flow Matching (CFM) (Section~\ref{sec:flow_matching_enhancement}), which introduces a margin-based repulsive objective to separate interwoven dynamics in the flow space (Figure~\ref{fig:2}c). 

\subsection{Recursive Consistent Action Flow (RCAF)}
\label{sec:rcaf}

Single-step action generation improves inference efficiency, but direct one-step approximation may fail to capture intermediate trajectory corrections in high-dimensional action spaces. This can introduce spatial truncation errors, especially for manipulation tasks that require accurate joint or end-effector motion. Motivated by recursive consistency learning~\cite{sunany}, we propose Recursive Consistent Action Flow (RCAF), a robotic action generation method that constructs a corrected velocity target by comparing a long-span teacher prediction with a recursively estimated multi-step trajectory.

\textbf{Time Span Sampling.} RCAF adopts a multi-timestep sampling strategy~\cite{frans2024one} by augmenting the policy network with a time span parameter $\Delta t$, denoted as $v_\theta(x_t, t, \Delta t, c)$. For each expert action sequence $x_1$ and its Gaussian noise $x_0 \sim \mathcal{N}(0, I)$, we sample a random span $\Delta t \sim \mathcal{U}[dt_{\text{min}}, 1.0]$ and a starting time $t_s \sim \mathcal{U}[0, 1.0 - \Delta t]$. Based on Eq.~\ref{eq:ot_path}, the noisy action state at $t_s$ is determined by $x_s = (1 - t_s)x_0 + t_s x_1$, while the OT target velocity driving this evolution remains constant as $v_{\text{true}} = x_1 - x_0$.

\textbf{Recursive Displacement Estimation.} To quantify the deviation from the curved manifold, an EMA teacher $\theta_{\text{EMA}}$ first executes a long-term prediction over $\Delta t$ to estimate the average velocity $v_{\text{long}} = v_{\theta_{\text{EMA}}}(x_s, t_s, \Delta t, c)$, yielding a single-step displacement $D_{\text{long}} = v_{\text{long}} \cdot \Delta t$. For a more precise trajectory evolution, RCAF divides $\Delta t$ into an Anchor Step ($dt_{\text{anc}} = \rho \cdot \Delta t$, with $\rho = 0.2$) and an $N$-step Recursive Tail. The anchor step strictly propagates the state using the OT target velocity to obtain the intermediate state $x_{\text{anchor}} = x_s + v_{\text{true}} \cdot dt_{\text{anc}}$ at $t_1 = t_s + dt_{\text{anc}}$. Starting from $x_{\text{anchor}}$, the teacher performs iterative integration over the remaining interval $[t_1, t_e]$ with sub-step size $dt_s = (t_e - t_1)/N$, accumulating the precise recursive displacement $D_{\text{rec}} = \sum_{i=1}^N v_{\theta_{\text{EMA}}}(x_i, t_i, dt_s, c) \cdot dt_s$.

\textbf{Residual Correction.} Given the temporal span $\Delta t$, a discrepancy $\Delta D = D_{\text{long}} - D_{\text{rec}}$ arises between the single-step linear leap and the manifold-adhering recursive prediction. We formulate this deviation as the accumulated spatial error of the long-term prediction. By amortizing $\Delta D$ over the anchor step, we derive the teacher model's implicit starting velocity estimate $v_{\text{est}}$ and the corresponding \textbf{directional residual correction} $\xi$:
\begin{equation}
\label{eq:xi_combined}
v_{\text{est}} = \frac{D_{\text{long}} - D_{\text{rec}}}{dt_{\text{anc}}}, \quad \xi = \operatorname{Clamp}(v_{\text{est}} - v_{\text{true}}, -\gamma, \gamma)
\end{equation}
where $\gamma$ bounds the correction magnitude to prevent gradient explosion. The intuition behind $\xi$ is to guide the student model—constrained to a linear projection—to adjust its initial heading to better align at the final state of the teacher's curved evolution. Thus, $\xi$ acts as an explicit error vector, quantifying the directional offset required to prevent the single-step model from diverging from the non-linear action manifold.

\textbf{Training Objective.} To achieve precise single-step inference, we align the student's predicted velocity $v_\theta$ with a residually corrected target $v_{\text{target}} = \text{sg}(v_{\text{long}} - \xi)$, where $\text{sg}(\cdot)$ denotes the Stop-Gradient operation to ensure target stability. The final RCAF loss function is formulated as:
\begin{equation}
\label{eq:loss_rcaf_final}
\mathcal{L}_{\operatorname{RCAF}}
=
\mathbb{E}_{x_0,x_1,c,t_s,\Delta t}
\left[
\left\|
v_{\theta}(x_s, t_s, \Delta t, c)
-
v_{\text{target}}
\right\|_2^2
\right],
\end{equation}
Through this explicit recursive correction, RCAF constructs an $N$-th order temporal consistency constraint. Mathematically, as derived in Appendix~\ref{sec:theoretical_analysis}, this target construction can be interpreted as an implicit high-gain feedback mechanism, where any subtle deviation from the curved manifold is inversely amplified to provide a corrective direction. This mechanism helps the single-step policy compensate for spatial truncation errors, providing a favorable trade-off between inference efficiency and manipulation fidelity. The complete procedure for generating the RCAF target is summarized in Algorithm~\ref{alg:rcaf}.

\begin{algorithm}[htbp]
\caption{Recursive Consistent Action Flow (RCAF)}
\label{alg:rcaf}
\begin{algorithmic}[1]
\REQUIRE Expert actions $x_1$, noise $x_0$, EMA teacher $v_{\theta_{\text{EMA}}}$, steps $N$, ratio $\rho$, condition $c$.
\STATE \textbf{Function} $\operatorname{RCAF}(x_1^C, x_0^C, v_{\theta_{\text{EMA}}}, c)$:
    \STATE Sample span $\Delta t \sim \mathcal{U}[dt_{\text{min}}, 1]$, start $t_s \sim \mathcal{U}[0, 1 - \Delta t]$, set $t_{\text{e}} = t_s + \Delta t$
    \STATE $x_{\text{s}} = (1 - t_s)x_0 + t_s x_1$, \quad $v_{\text{true}} = x_1 - x_0$
    
    \STATE \textcolor{gray}{\# 1. Teacher long-term prediction }
    \STATE $v_{\text{long}} = v_{\theta_{\text{EMA}}}(x_{\text{s}}, t_s, \Delta t, c)$
    
    \STATE \textcolor{gray}{\# 2. Anchor step via OT target velocity}
    \STATE $dt_{\text{anc}} = \rho \cdot \Delta t$, \quad $t_1 = t_s + dt_{\text{anc}}$, \quad $x_{\text{curr}} = x_{\text{s}} + v_{\text{true}} \cdot dt_{\text{anc}}$
    
    \STATE \textcolor{gray}{\# 3. Teacher N-step recursive tail estimation}
    \STATE $D_{\text{rec}} \leftarrow 0$, \quad $dt_s = (t_{\text{e}} - t_1)/N$
    \FOR{$i = 1$ \TO $N$}
        \STATE $t_c = t_1 + (i-1)dt_s$, $v_i = v_{\theta_{\text{EMA}}}(x_{\text{curr}}, t_c, dt_s, c)$
        \STATE $D_{\text{rec}} \leftarrow D_{\text{rec}} + v_i \cdot dt_s$, \quad $x_{\text{curr}} \leftarrow x_{\text{curr}} + v_i \cdot dt_s$
    \ENDFOR
    
    \STATE \textcolor{gray}{\# 4. Residual correction $\xi$ and target computing}
    \STATE $\xi = \operatorname{Clamp}\left( \frac{v_{\text{long}} \cdot \Delta t - D_{\text{rec}}}{dt_{\text{anc}}} - v_{\text{true}}, \ -\gamma, \ \gamma \right)$
    \STATE $v_{\text{target}} = \operatorname{sg}(v_{\text{long}} - \xi)$
    
    \RETURN $x_{\text{s}}, t_s, \Delta t, v_{\text{target}}$
\end{algorithmic}
\end{algorithm}

\subsection{Dual-Timestep Frequency Consistency (DTFC)}
\label{sec:dtfc}

Although RCAF mitigates spatial truncation errors, optimizing the single-step model solely with global MSE can bias predictions toward low-frequency, overly smooth motions. This may suppress high-frequency manipulation details that are important for precise control, such as fine fingertip adjustments and rapid contact-rich corrections.

To overcome this bottleneck, inspired by~\cite{su2025freqpolicy}, we propose the Dual-Timestep Frequency Consistency (DTFC) mechanism. DTFC encourages adaptive spectral consistency across different flow timesteps, preventing high-frequency manipulation details from being overly suppressed during one-step generation.

To leverage the temporal dynamics of action chunks, we employ the 1D Type-II Discrete Cosine Transform ($\mathcal{DCT}$) to map the time-domain velocity predictions $v \in \mathbb{R}^{H \times D}$ into the spectral domain. This transformation separates high- and low-frequency signals across the prediction horizon $H$, which often provides a compact representation with dominant low-frequency components. Specifically, we independently sample two flow timesteps $t_a, t_b \sim \mathcal{U}[0, 1]$ and construct their corresponding noisy states $x_a$ and $x_b$ via the standard Optimal Transport formulation (Eq.~\ref{eq:ot_path}). Next, we predict the corresponding action flow velocities and apply the $\mathcal{DCT}$ operator to map these sequential features into the spectral domain:
\begin{equation}
\label{eq:v_freq}
\hat{v}_a = \mathcal{DCT}(v_\theta(x_a, t_a, 0, c)), \quad \hat{v}_b = \mathcal{DCT}(v_\theta(x_b, t_b, 0, c))
\end{equation}
where $\hat{v}_a,\hat{v}_b\in\mathbb{R}^{H\times D}$ denote the frequency-domain representations of the predicted velocities. 
The span input is set to $0$ to indicate the standard instantaneous velocity prediction branch. 
To quantify spectral consistency across different flow timesteps, we compute the discrepancy at frequency band $k$ as
\begin{equation}
E_{\text{abs}}^{(k)}
=
\frac{1}{D}
\sum_{d=1}^{D}
\left|
\hat{v}_a^{(k,d)}-\hat{v}_b^{(k,d)}
\right|.
\end{equation}

To prevent the model from merely fitting easy-to-learn low-frequency components, we design a combined adaptive weight $\omega_{\text{comb}}^{(k)}$ for the $k$-th frequency band by coupling two core mechanisms:

\textbf{(1) Dynamic Focal Weight:} This mechanism allocates attention based on the relative spectral error, encouraging the model to actively focus on the frequencies with the largest discrepancies. It is defined as $\omega_{\text{focal}}^{(k)} = \left( 1 + E_{\text{abs}}^{(k)} / (\max_j E_{\text{abs}}^{(j)} + \epsilon) \right)^\alpha$, where $\alpha \ge 0$ controls the penalty degree for hard samples.

\textbf{(2) Dual-Timestep Perception Mask:} We further introduce a timestep-aware spectral mask to adapt the frequency weights according to the sampled flow timesteps. 
Specifically, we normalize the discrete frequency index into $f_k\in[0,1]$ and use it as a frequency coordinate. This design provides a simple spectral prior: earlier flow stages are assigned larger weights on lower-frequency bands to emphasize global action structure, while later stages place more weight on higher-frequency bands to preserve local manipulation details. Accordingly, we construct a Gaussian mask centered around the sampled timesteps $t_a$ and $t_b$:
\begin{equation}
\label{eq:w_time}
\omega_{\text{time}}^{(k)} = \max_{t \in \{t_a, t_b\}} \exp\left(-\frac{(f_k - t)^2}{2\sigma^2}\right)
\end{equation}

We fuse these weights via element-wise multiplication ($\omega_{\text{comb}}^{(k)} = \omega_{\text{focal}}^{(k)} \cdot \omega_{\text{time}}^{(k)}$) and channel normalization. The final DTFC loss penalizes spectral inconsistencies as:
\begin{equation}
\label{eq:loss_dtfc}
\mathcal{L}_{\operatorname{DTFC}}
=
\mathbb{E}_{x_0,x_1,c,t_a,t_b}
\left[
\sum_{k=1}^{H}
\omega_{\text{comb}}^{(k)}
\cdot
\frac{1}{D}
\sum_{d=1}^{D}
\left(
\hat{v}_a^{(k,d)}-\hat{v}_b^{(k,d)}
\right)^2
\right].
\end{equation}

\begin{algorithm}[htbp]
\caption{Dual-Timestep Frequency Consistency (DTFC)}
\label{alg:dtfc_short}
\begin{algorithmic}[1]
\REQUIRE State $x_a$, time $t_a$, pred. vel. $v_a$, target vel. $v_{\text{true}}$, student $v_\theta$, condition $c$, hyperparams $\sigma, \alpha, \epsilon$

\STATE \textbf{Function} $\operatorname{DTFC}(x_a, t_a, v_a, v_{\text{true}}, v_\theta, c)$:
    \STATE \textcolor{gray}{\# Sample random timestep and compute state}
    \STATE Sample $t_b \sim \mathcal{U}[0, 1]$; compute state $x_b = x_a + (t_b - t_a) v_{\text{true}}$
    
    \STATE \textcolor{gray}{\# Predict velocity and transform to frequency}
    \STATE Predict velocity at $t_b$: $v_b = v_\theta(x_b, t_b, 0, c)$
    \STATE $\hat{v}_a, \hat{v}_b = \operatorname{DCT}(v_a), \operatorname{DCT}(v_b)$
    \STATE Let $E_{\text{abs}}^{(k)}=\frac{1}{D}\sum_{d=1}^{D}\left|\hat{v}_a^{(k,d)}-\hat{v}_b^{(k,d)}\right|$ denote the absolute spectral error

    \STATE \textcolor{gray}{\# 1. Adaptive Focal Weight}
    \STATE $\omega_{\text{focal}} = \Big(1 + \frac{E_{\text{abs}}}{\max(E_{\text{abs}}) + \epsilon}\Big)^\alpha$
    
    \STATE \textcolor{gray}{\# 2. Dual-Timestep Gaussian Mask}
    \STATE $\omega_{\text{time}} = \max_{t \in \{t_a, t_b\}} \exp\Big(-\frac{(f - t)^2}{2\sigma^2}\Big)$
    
    \STATE \textcolor{gray}{\# 3. Aggregate weights and compute final loss}
    \STATE $\omega_{\text{combined}} = \omega_{\text{focal}} \odot \omega_{\text{time}}$ \quad \algorithmiccomment{\textit{$\odot$ is element-wise product}}
    \RETURN $\operatorname{Mean}\Big( \frac{\omega_{\text{combined}}}{\operatorname{Mean}(\omega_{\text{combined}})} \odot (E_{\text{abs}})^2 \Big)$
\end{algorithmic}
\end{algorithm}

\textbf{Frequency Curriculum Learning.} To improve training stability and prevent the model from being prematurely overwhelmed by high-frequency penalty gradients before the foundational spatial flow field converges, we introduce a linear warmup curriculum. Specifically, the DTFC objective is dynamically scaled by a curriculum coefficient $\alpha_{\text{curr}} = \min(1.0, \text{epoch} / E_{\text{warmup}})$, where $E_{\text{warmup}}$ denotes the warmup duration. This dynamic weighting ensures a smooth training transition from learning macroscopic spatial trajectories to refining fine-grained high-frequency manipulation details.

\vspace{-1mm}
\subsection{Contrastive Flow Matching (CFM)}
\label{sec:flow_matching_enhancement}
In standard flow matching, single-step models can suffer from mode averaging when different valid actions correspond to similar observations or diverse spatial layouts. As illustrated in Figure~\ref{fig:2}c, trajectories generated from 100 Gaussian noise samples across two distinct scenes show noticeable overlap in the flow space, indicating entangled action modes. Such entanglement can lead to two issues. 
First, it may produce ambiguous actions when different modes require conflicting motion directions. Second, interwoven flow fields can increase trajectory curvature, making one-step approximation more prone to spatial truncation errors.

To reduce multimodal flow entanglement, we introduce Contrastive Flow Matching (CFM), a margin-based objective that encourages non-matching action flows to remain separated in the velocity space, thereby reducing mode averaging. The contrastive loss is defined as:
\begin{equation}
\label{eq:loss_cfm}
\mathcal{L}_{\operatorname{CFM}}
=
\mathbb{E}_{x_0,x_1,c,t}
\left[
\mathcal{L}_{\text{pos}}
+
\lambda_{\operatorname{CFM}}
\cdot
\max(0, m-\mathcal{L}_{\text{neg}})
\right],
\end{equation}

where $\mathcal{L}_{\text{pos}} = \|v_\theta(x_t, t, 0, c) - v_{\text{target}}\|_2^2$ minimizes the distance between the predicted flow and the target velocity under the current condition $c$. Conversely, the negative distance $\mathcal{L}_{\text{neg}} = \|v_\theta(x_t, t, 0, c) - v_{\text{target}}^{\text{neg}}\|_2^2$ calculates the repulsive distance from the target velocity fields of non-matching scenes, where $m$ denotes the margin threshold.

\textbf{In-Batch Cyclic Negative Sampling.} To minimize the computational overhead of generating negative samples, we adopt an efficient in-batch cyclic shifting strategy inspired by~\cite{stoica2025contrastive}. Within each training batch, the target velocity field is shifted along the batch dimension:
$v_{\text{target}}^{\text{neg}} = \operatorname{Roll}(v_{\text{target}}, \text{shift}=1)$.
This simple strategy provides non-matching target flows with negligible computational overhead. As shown in Figure~\ref{fig:2}c, the resulting repulsive objective encourages different action modes to become more separated in the flow space. The complete procedure of the proposed CFM objective is summarized in Algorithm~\ref{alg:cfm}.

\begin{algorithm}[htbp]
\caption{Contrastive Flow Matching (CFM)}
\label{alg:cfm}
\begin{algorithmic}[1]
\REQUIRE Predicted velocity $v_{\text{pred}}$, target velocity $v_{\text{target}}$, margin threshold $m$, penalty weight $\lambda_{\operatorname{CFM}}$

\STATE \textbf{Function} $\operatorname{CFM}(v_{\text{pred}}, v_{\text{target}})$:
    \STATE \textcolor{gray}{\# 1. Compute flow matching loss for positive pairs}
    \STATE $\mathcal{L}_{\text{pos}} = \| v_{\text{pred}} - v_{\text{target}} \|_2^2$ 
    
    \STATE \textcolor{gray}{\# 2. Generate negative targets via in-batch cyclic shifting}
    \STATE $v_{\text{target}}^{\text{neg}} = \operatorname{Roll}(v_{\text{target}}, \text{shift}=1)$
    
    \STATE \textcolor{gray}{\# 3. Compute distance to negative samples}
    \STATE $\mathcal{L}_{\text{neg}} = \| v_{\text{pred}} - v_{\text{target}}^{\text{neg}} \|_2^2$
    
    \STATE \textcolor{gray}{\# 4. Compute contrastive regularization with margin}
    \STATE $\mathcal{L}_{\text{contrastive}} = \max(0, m - \mathcal{L}_{\text{neg}})$
    
    \STATE \textcolor{gray}{\# 5. Aggregate the final CFM loss}
    \RETURN $\mathcal{L}_{\text{pos}} + \lambda_{\operatorname{CFM}} \cdot \mathcal{L}_{\text{contrastive}}$
\end{algorithmic}
\end{algorithm}

\subsection{Overall Training Objective}

By integrating the three core modules detailed above, our end-to-end joint training objective is formulated as follows:
\begin{equation}
\label{eq:loss_total}
\mathcal{L}_{\text{Total}} = \mathcal{L}_{\operatorname{RCAF}} + \mathcal{L}_{\operatorname{CFM}} + \lambda_{\text{freq}} \cdot \alpha_{\text{curr}} \cdot \mathcal{L}_{\operatorname{DTFC}}
\end{equation}
where $\lambda_{\text{freq}}$ serves as the base balancing coefficient for the frequency consistency loss, and $\alpha_{\text{curr}}$ is the epoch-dependent curriculum multiplier (defined in Section~\ref{sec:dtfc}). 

By jointly optimizing $\mathcal{L}_{\text{Total}}$, the policy is regularized through recursive correction (RCAF), frequency consistency (DTFC), and contrastive flow matching (CFM). These objectives jointly improve the fidelity of single-step action generation while preserving the efficiency of 1-NFE inference. The overall training pipeline of our framework is summarized in Algorithm~\ref{alg:overall}. 

\begin{algorithm}[htbp]
\caption{Overall Training Pipeline for our Framework}
\label{alg:overall}
\begin{algorithmic}[1]
\REQUIRE Expert dataset $\mathcal{D}$, student model $v_\theta$, EMA teacher $v_{\theta_{\text{EMA}}}$, flow batch ratio $r_{\text{flow}}$.
\WHILE{not converged}
    \STATE Sample expert actions $x_1 \sim \mathcal{D}$ and noise $x_0 \sim \mathcal{N}(0, \mathbf{I})$
    \STATE Partition batch into flow-matching pair $(x_1^F, x_0^F)$ and consistency pair $(x_1^C, x_0^C)$ via ratio $r_{\text{flow}}$
    \STATE Sample $t_a \sim \mathcal{U}[0,1]$ and set 
    $x_{t_a}^F=(1-t_a)x_0^F+t_a x_1^F$
    \STATE $v_{\text{true}}^F=x_1^F-x_0^F$, \quad
    $v_a^F=v_\theta(x_{t_a}^F,t_a,0,c)$
    \STATE $\mathcal{L}_{\operatorname{CFM}}=\operatorname{CFM}(v_a^F,v_{\text{true}}^F)$
    \STATE $\mathcal{L}_{\operatorname{DTFC}}=
    \operatorname{DTFC}(x_{t_a}^F,t_a,v_a^F,v_{\text{true}}^F,v_\theta,c)$
    \STATE $x_{\text{start}}^C, t_s, \Delta t, v_{\text{target}}^C = \operatorname{RCAF}(x_1^C, x_0^C, v_{\theta_{\text{EMA}}}, c)$ 
    \STATE $\mathcal{L}_{\operatorname{RCAF}} = \| v_\theta(x_{\text{start}}^C, t_s, \Delta t, c) - v_{\text{target}}^C \|_2^2$
    \STATE $\mathcal{L}_{\text{Total}} = \mathcal{L}_{\operatorname{RCAF}} + \mathcal{L}_{\operatorname{CFM}} + \lambda_{\text{freq}} \alpha_{\text{curr}} \mathcal{L}_{\operatorname{DTFC}}$
    \STATE $\theta \leftarrow \theta - \eta \nabla_\theta \mathcal{L}_{\text{Total}}$
    \STATE $\theta_{\text{EMA}} \leftarrow \beta \theta_{\text{EMA}} + (1-\beta)\theta$ \hfill \textcolor{gray}{\# EMA update}
    
\ENDWHILE
\end{algorithmic}
\end{algorithm}

\vspace{-3mm}
\subsection{Single-Step Inference}
\label{sec:inference}

Given a condition $c$ extracted from current observations, we initialize a pure Gaussian noise trajectory $x_0 \sim \mathcal{N}(0, \mathbf{I})$. For single-step inference ($N=1$), we fix the starting time at $t=0$ and the integration span at $\Delta t = 1.0$. The policy network $v_\theta$ directly maps the noise $x_0$ to the global action velocity $v_{\text{pred}}$. The final physical action trajectory $x_1$ is then generated via a single Euler integration step: $x_1 = x_0 + v_\theta(x_0, t=0, \Delta t=1.0, c)$. This non-iterative paradigm avoids iterative ODE solving, enabling low-latency closed-loop control while maintaining competitive manipulation performance.


\section{Experiments}
\label{sec:experiments}

We evaluate the proposed framework on 33 tasks across four simulation benchmarks using both 2D image and 3D point-cloud inputs. We further conduct ablation studies and real-world robot experiments to analyze the effectiveness and deployability of our method.

\begin{figure*}[htbp]
    \centering
    \includegraphics[width=0.9\textwidth]{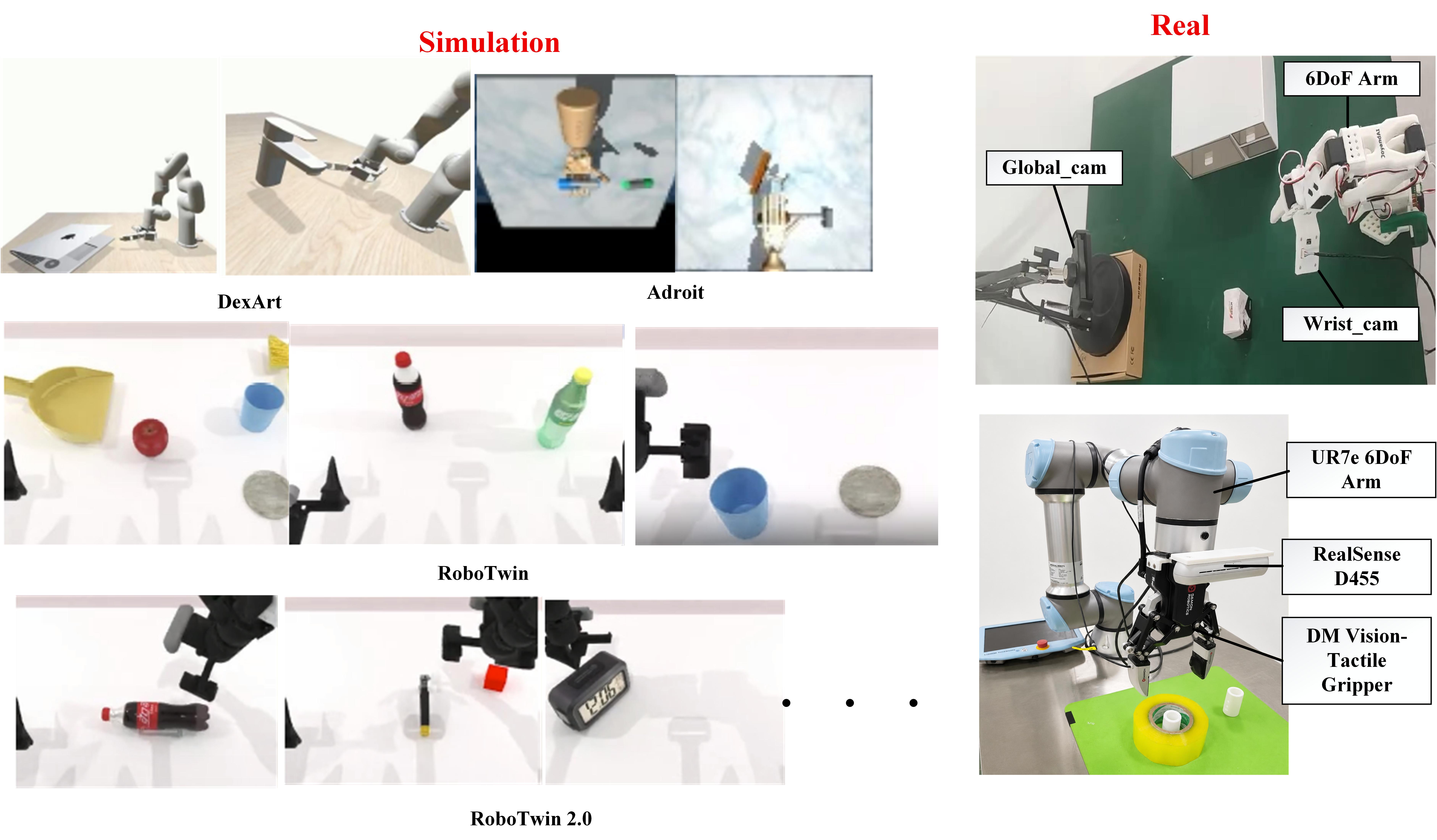}
    \caption{
        \textbf{Overview of simulation benchmarks and real-world experiments.}
        Left: the simulation evaluation suite includes 33 diverse tasks across DexArt, Adroit, RoboTwin 1.0, and RoboTwin 2.0, covering dexterous manipulation and complex bimanual coordination.
        Right: the real-world evaluation consists of multiple experiments on two physical 6-DoF robotic platforms: the SO101 arm with a global camera and a wrist-mounted camera, and the UR7E arm equipped with a RealSense D455 camera and a DM vision-tactile gripper.
    }
    \label{fig:benchmark_tasks}
\end{figure*}

\subsection{Simulation Benchmarks}
As shown in Figure~\ref{fig:benchmark_tasks}, we evaluate our method under two observation modalities.
\textbf{2D Visuomotor Control.} 
Following established protocols~\cite{chi2025diffusion,yan2025maniflow}, we use a pre-trained ResNet-18~\cite{he2016deep} to encode image observations. The 2D evaluation includes 5 bimanual manipulation tasks from RoboTwin 1.0~\cite{mu2025robotwin}, 21 cross-domain tasks from RoboTwin 2.0~\cite{chen2025robotwin} for OOD generalization, and 7 dexterous manipulation tasks from Adroit~\cite{rajeswaran2017learning} and DexArt~\cite{bao2023dexart}.

\textbf{3D Visuomotor Control.} 
To evaluate 3D spatial reasoning, we extract geometric features from point clouds and adopt 3D-conditioned policy architectures following~\cite{ze20243d,yan2025maniflow}. The 3D setting is evaluated on RoboTwin 1.0, Adroit, and DexArt to test the policy's ability to process raw spatial geometry.

\subsection{Evaluation Metrics and Implementation Details}

\textbf{Evaluation Metrics.} 
We report the task \textit{Success Rate}, the \textit{Number of Function Evaluations (NFE)}, and \textit{Cross-Domain Generalization} performance. All models are trained and evaluated under the same computational settings. Our method uses only \textbf{1 NFE} for single-step generation, while strong generative baselines such as 3D Diffusion Policy~\cite{ze20243d} and ManiFlow~\cite{yan2025maniflow} use their default 10-step inference configuration.

\textbf{Implementation Details.} 
Detailed hyperparameters, temporal horizon settings, and module-specific configurations for RCAF, DTFC, and CFM are provided in the APPENDIX A.

\subsection{Comparisons with the State-of-the-Art}

As shown in Table~\ref{tab:compact}, we compare our 1-NFE framework with representative generative visuomotor policies across both 2D image-based and 3D point-cloud settings. Across all benchmarks, our method achieves the best overall average performance while requiring only a single function evaluation, whereas strong generative baselines typically use 10 inference steps.

\textbf{2D Visuomotor Setting.} In the image-based setting, our method achieves an overall average success rate of $65.3\%$, outperforming 10-step ManiFlow ($56.5\%$) and Diffusion Policy ($39.4\%$) while using only 1 NFE. The improvement is particularly clear on RoboTwin, where our method increases the average success rate from $46.1\%$ to $58.9\%$ compared with ManiFlow. On challenging spatial reasoning tasks such as \textit{Pick Apple Messy}, \textit{Diverse Bottles Pick}, and \textit{Empty Cup Place}, our method consistently improves over the strongest baseline, suggesting that the proposed one-step policy better preserves action-flow geometry under complex visual layouts. On Adroit and DexArt, our method also achieves higher average performance, indicating that the proposed modules remain effective in dexterous manipulation scenarios with high-dimensional actions.

\textbf{3D Point Cloud Setting.} In the 3D setting, our method achieves an overall average success rate of $74.2\%$, improving over 3D ManiFlow ($66.5\%$) under the same point-cloud observation modality. The gains are especially notable on Adroit, where the average success rate improves from $78.6\%$ to $88.9\%$. For example, on the \textit{pen} task, our method substantially improves performance over 3D ManiFlow, suggesting better handling of fine-grained dexterous motion. On RoboTwin and DexArt, our method also improves the average success rate from $61.9\%$ to $67.2\%$ and from $63.2\%$ to $66.5\%$, respectively. These results indicate that the proposed framework can benefit both image-based and geometry-based visuomotor policies.

\textbf{Cross-Domain Generalization.} To evaluate robustness in unseen environments, we further conduct cross-domain experiments on RoboTwin 2.0 and compare our single-step policy with ManiFlow, a strong generative policy baseline (Table~\ref{tab:maniflow_ours}). Despite using only one function evaluation, our method improves the average success rate from $28.8\%$ to $34.5\%$. The gains are consistent across most tasks, including \textit{Click Alarmclock}, \textit{Dump Bin Bigbin}, \textit{Shake Bottle Horiz.}, and \textit{Turn Switch}. These improvements suggest that the proposed RCAF, DTFC, and CFM modules help reduce the approximation gap of single-step generation, leading to better robustness under visual and physical domain shifts.

Overall, the results show that our method achieves competitive or superior manipulation performance compared with strong 10-step generative policy baselines, while reducing the inference process to a single forward pass. This demonstrates that high-fidelity one-step visuomotor generation is feasible when spatial correction, frequency preservation, and multimodal flow separation are jointly considered.

\begin{table*}[t]
\centering
\caption{Main simulation results across dexterous manipulation and bimanual manipulation benchmarks.}
\vspace{2mm} 

\label{tab:compact}
\resizebox{\textwidth}{!}{
\begin{tabular}{l | c | c | cccc | ccccc}
\toprule
\multirow{2}{*}{Algorithm $\backslash$ Task} & \multirow{2}{*}{Obs.} & \multirow{2}{*}{NFE} & \multicolumn{4}{c|}{Adroit (10 demos)} & \multicolumn{5}{c}{DexArt (100 demos)} \\
\cmidrule(lr){4-7} \cmidrule(lr){8-12}
& & & hammer & door & pen & Average & laptop & faucet & bucket & toilet & Average \\
\midrule
Diffusion Policy & Img & 10 & $54.0\pm3.6$ & $41.8\pm2.7$ & $18.5\pm2.5$ & $38.1\pm2.9$ & $81.7\pm2.1$ & $29.3\pm2.1$ & $26.0\pm2.4$ & $77.3\pm1.9$ & $53.6\pm2.1$ \\
Flow Matching Policy & Img & 10 & $55.7\pm4.2$ & $40.0\pm1.6$ & $21.2\pm0.8$ & $39.0\pm2.2$ & $81.7\pm2.5$ & $31.3\pm3.7$ & $24.0\pm2.2$ & $76.3\pm1.2$ & $53.3\pm2.4$ \\
2D ManiFlow & Img & 10 & $\mathbf{100.0}\pm0.0$ & $67.0\pm2.2$ & $ 56.0\pm3.6$ & $ 74.3\pm1.9$ & $ 85.7\pm2.1$ & $32.3\pm0.5$ & $29.7\pm3.4$ & $77.7\pm3.3$ & $56.3\pm2.3$ \\
\rowcolor{gray!15} \textbf{Ours} & Img & \textbf{1} & $\mathbf{100.0\pm0.0}$ & $\mathbf{68.0\pm2.7}$ & $\mathbf{59.0\pm1.4}$ & $\mathbf{75.6\pm2.1}$ & $\mathbf{88.0 \pm2.7}$ & $\mathbf{41.0\pm2.2}$ & $\mathbf{34.0\pm2.2}$ & $\mathbf{82.0\pm2.7}$ & $\mathbf{61.3\pm2.4}$ \\
\midrule
3D Diffusion Policy & PC & 10 & $100.0\pm0.0$ & $76.7\pm4.7$ & $ 56.7\pm2.6 $ & $77.8\pm2.4$ & $89.7\pm0.9$ & $41.7\pm0.5$ & $31.3\pm0.5$ & $ 79.7\pm0.9 $ & $60.6\pm0.7$ \\
3D Flow Matching* & PC & 10 & $100.0\pm0.0$ & $77.7\pm6.1$ & $53.5\pm3.9$ & $77.1\pm3.3$ & $92.7\pm1.2$ & $42.0\pm0.8$ & $32.3\pm1.9$ & $ 79.7\pm0.5$ & $61.7\pm1.1$ \\
3D ManiFlow & PC & 10 & $ 100.0\pm0.0$ & $ 80.3\pm1.2$ & $55.5\pm5.8$ & $ 78.6\pm2.3 $ & $ 93.0\pm1.6 $ & $ \mathbf{45.0\pm3.6} $ & $ 35.3\pm2.1 $ & $79.3\pm3.3$ & $ 63.2\pm2.7 $ \\
\rowcolor{gray!15} \textbf{Ours} & PC & \textbf{1} & $\mathbf{100.0\pm0.0}$ & $\mathbf{85.0\pm0}$ & $\mathbf{81.7\pm2.9}$ & $\mathbf{88.9\pm1.0}$ & $\mathbf{96.0\pm2.2}$ & $44.0\pm2.2 $ & $\mathbf{39.2\pm2.0}$ & $\mathbf{86.7\pm2.9}$ & $\mathbf{66.5\pm2.2}$ \\
\bottomrule
\end{tabular}
} 

\vspace{0.5mm} 

\resizebox{\textwidth}{!}{
\begin{tabular}{l | c | c | cccccc | c}
\toprule
\multirow{2}{*}{Algorithm $\backslash$ Task} & \multirow{2}{*}{Obs.} & \multirow{2}{*}{NFE} & \multicolumn{6}{c|}{RoboTwin (50 demos)} & \multirow{2}{*}{\textbf{Overall Avg.}} \\

\cmidrule(lr){4-9}
& & & Pick Apple Messy & Diverse Bottles Pick & Dual Bottles Pick Hard & Empty Cup Place & Shoe Place & Average & \\
\midrule
Diffusion Policy & Img & 10 & $17.0\pm0.8$ & $36.3\pm2.4$ & $41.3\pm3.7$ & $42.0\pm1.6$ & $7.3\pm2.9$ & $28.8\pm2.3$ & $39.4\pm2.3$ \\
Flow Matching Policy & Img & 10 & $15.3\pm1.9$ & $32.0\pm4.5$ & $43.0\pm0.0$ & $38.0\pm5.4$ & $7.3\pm1.7$ & $27.1\pm2.7$ & $38.8\pm2.5$ \\
2D ManiFlow & Img & 10 & $ 37.3\pm4.8$ & $ 37.0\pm1.6 $ & $ 47.3\pm2.1 $ & $ 63.7\pm1.2$ & $ 45.3\pm3.7 $ & $ 46.1\pm2.7 $ & $ 56.5\pm2.4 $ \\
\rowcolor{gray!15} \textbf{Ours} & Img & \textbf{1} & $\mathbf{64.3\pm0.6}$ & $\mathbf{50.0\pm1.0}$ & $\mathbf{53.7\pm1.2}$ & $ \mathbf{76.7\pm0.6} $ & $\mathbf{49.7\pm1.2}$ & $\mathbf{58.9\pm0.9}$ & $\mathbf{65.3\pm1.8}$ \\
\midrule
3D Diffusion Policy & PC & 10 & $9.3\pm3.7$ & $38.3\pm7.1$ & $46.3\pm2.5$ & $73.0\pm0.8$ & $46.5\pm2.5$ & $42.7\pm3.3$ & $57.4\pm2.2$ \\
3D Flow Matching* & PC & 10 & $16.0\pm7.1$ & $56.3\pm6.6$ & $46.5\pm0.5$ & $ 82.3\pm1.7 $ & $39.3\pm15.5$ & $48.1\pm6.3$ & $59.9\pm2.8$ \\
3D ManiFlow & PC & 10 & $ 42.0\pm0.8$ & $ 72.3\pm1.7$ & $ 54.0\pm2.2 $ & $72.7\pm4.8$ & $ \mathbf{68.3\pm2.9} $ & $ 61.9\pm2.5 $ & $ 66.5\pm2.5 $ \\
\rowcolor{gray!15} \textbf{Ours} & PC & \textbf{1} & $\mathbf{46.3\pm3.2}$ & $\mathbf{79.0\pm1.7}$ & $\mathbf{58.0\pm2.6}$ & $ \mathbf{86.3\pm1.5} $ & $66.3\pm3.2 $ & $\mathbf{67.2\pm2.4}$ & $\mathbf{74.2\pm1.9}$ \\
\bottomrule
\end{tabular}
} 
\end{table*}

\begin{figure}[t]
    \centering
    \includegraphics[width=\columnwidth]{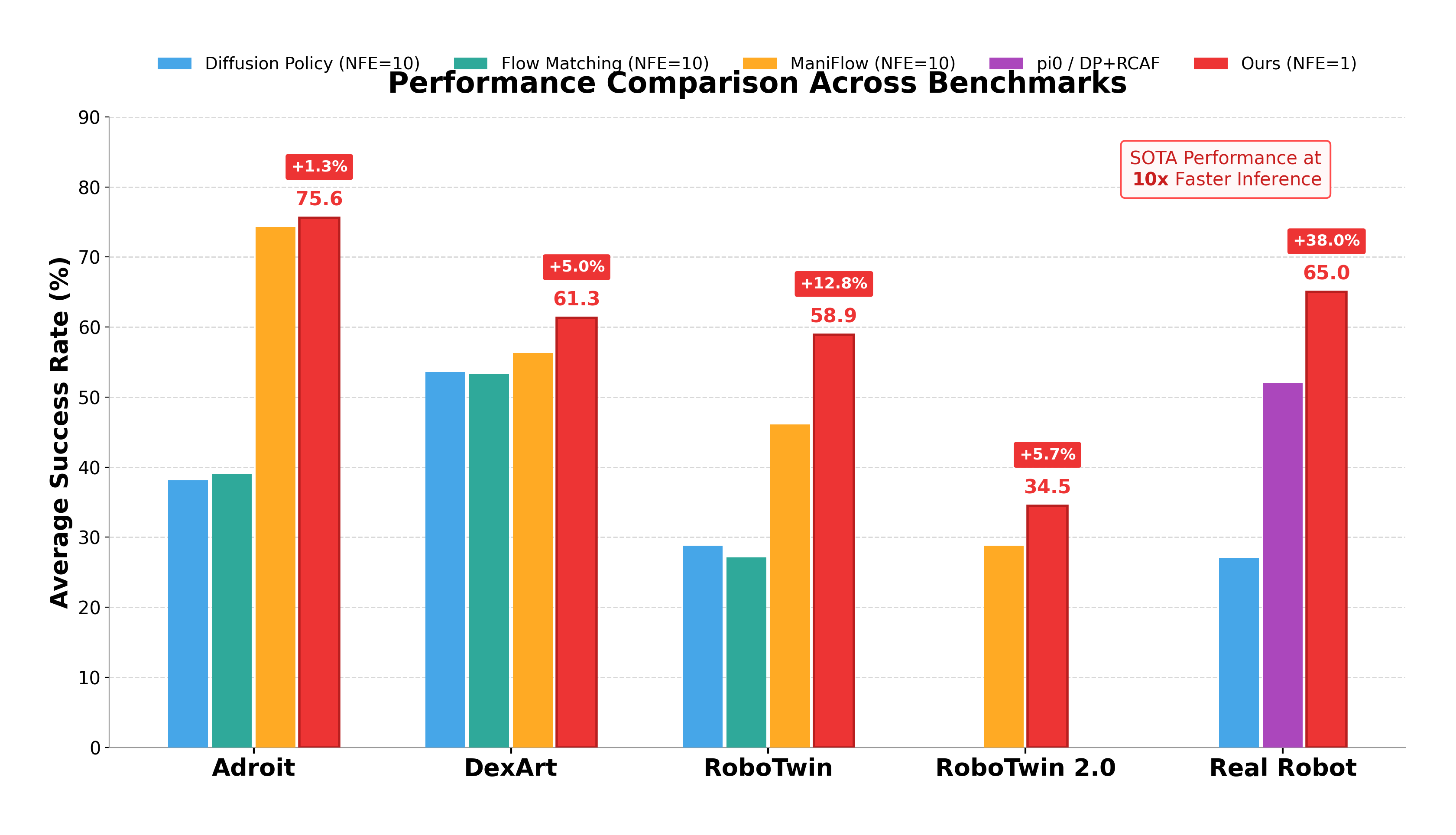}
    \vspace{-0.2cm}
    \caption{
        \textbf{Performance Comparison with SOTA Methods.} Average success rates across five image-based manipulation benchmarks. Our single-step framework (NFE=1) achieves strong average performance while requiring an order of magnitude fewer inference steps than multi-step generative baselines.
    }
    \label{fig:performance_comparison}
    \vspace{-0.4cm}
\end{figure}

\begin{table}[t]
\centering
\caption{Cross-domain performance comparison between ManiFlow and ours on RoboTwin 2.0. Best results are in bold.}
\label{tab:maniflow_ours}
\scriptsize
\setlength{\tabcolsep}{3pt}
\begin{tabular}{@{}lcc|lcc@{}}
\toprule
Task & ManiFlow & Ours & Task & ManiFlow & Ours \\
\midrule
Adjust Bottle & 84 & \textbf{88} 
& Pick Dual Bottles & 15 & \textbf{19} \\

Beat Block Hammer & 36 & \textbf{39} 
& Place Bread Basket & \textbf{6} & \textbf{6} \\

Click Alarmclock & 41 & \textbf{49} 
& Place Burger Fries & 21 & \textbf{31} \\

Click Bell & \textbf{22} & 20 
& Place Empty Cup & 3 & \textbf{8} \\

Dump Bin Bigbin & 63 & \textbf{71} 
& Press Stapler & 17 & \textbf{27} \\

Handover Block & 24 & \textbf{31} 
& Put Object Cabinet & 17 & \textbf{18} \\

Lift Pot & \textbf{32} & 31 
& Shake Bottle Horiz. & 36 & \textbf{50} \\

Move Can Pot & 39 & \textbf{45} 
& Stack Bowls Three & 48 & \textbf{51} \\

Move Playingcard Away & 33 & \textbf{41} 
& Stamp Seal & 3 & \textbf{7} \\

Open Laptop & 43 & \textbf{52} 
& Turn Switch & 17 & \textbf{31} \\

Pick Diverse Bottles & 4 & \textbf{10} 
& Average & 28.8 & \textbf{34.5} \\
\bottomrule
\end{tabular}
\vspace{-2mm}
\end{table}

\subsection{Ablation Studies}
\label{sec:ablation}

To analyze the contribution of each module, we conduct systematic ablation studies on the bimanual manipulation benchmark RoboTwin~\cite{mu2025robotwin}.

\textbf{Efficacy of Core Components.} Table~\ref{tab:ablation_modules} shows the incremental contribution of each module. The 10-step baseline~\cite{yan2025maniflow} yields a $46.1\%$ success rate. Introducing RCAF compresses inference to a single step and improves the success rate to $54.8\%$, indicating the effectiveness of its spatial truncation correction. Adding DTFC to preserve high-frequency manipulation details further raises success to $57.6\%$. Finally, integrating CFM to separate entangled action flows achieves the best performance of $59.8\%$.

\begin{table}[htbp]
\centering
\caption{Ablation of core framework components.}
\label{tab:ablation_modules}
\resizebox{\linewidth}{!}{
\begin{tabular}{@{} l ccccccc @{}}
\toprule
\textbf{Method} & \textbf{NFE} & \textbf{Pick} & \textbf{Diverse} & \textbf{Dual} & \textbf{Empty} & \textbf{Shoe} & \textbf{Average} \\
\midrule
Baseline & 10 & 37.3 & 37.0 & 47.3 & 63.7 & 45.3 & 46.1 \\
+ RCAF & 1 & 58.0 & 47.0 & 51.0 & 72.0 & 46.0 & 54.8 \\
+ RCAF + DTFC & 1 & 62.0 & 50.0 & 53.0 & 76.0 & 47.0 & 57.6 \\
\rowcolor{gray!15} \textbf{+ RCAF + DTFC + CFM} & \textbf{1} & \textbf{65.0} & \textbf{51.0} & \textbf{55.0} & \textbf{77.0} & \textbf{51.0} & \textbf{59.8} \\
\bottomrule
\end{tabular}
}
\end{table}

\textbf{RCAF Hyperparameters.} 
Table~\ref{tab:rcaf_ablation} studies the effects of the recursive step number $N$ and the anchor ratio $\rho$. When the anchor ratio becomes too small ($\rho \to 0$), the anchor step $dt_{\text{anc}}=\rho\Delta t$ approaches zero, making the residual correction in Eq.~\ref{eq:xi_combined} numerically unstable and leading to model divergence. Increasing the recursive order to $N=4$ brings only a marginal improvement over our default setting ($60.0\%$ vs. $59.8\%$), while requiring more recursive teacher evaluations during training. Considering stability, computational overhead, and performance, we use $N=2$ and $\rho=0.2$ as the default setting.

\begin{table}[htbp]
\centering
\caption{Ablation of RCAF hyperparameters ($N$ and $\rho$).}
\label{tab:rcaf_ablation}
\resizebox{\columnwidth}{!}{
\begin{tabular}{l | ccccc | c}
\toprule
\textbf{Settings} & Pick & Diverse & Dual & Empty & Shoe & \textbf{Average} \\
\midrule
$\rho=0.2, N=2$ (Ours) & 65 & 51 & 55 & 77 & 51 & 59.8 \\
$\rho=0.2, N=3$ & 59 & 48 & 52 & 76 & 50 & 57.0 \\
$\rho=0.2, N=4$ & 64 & 52 & 54 & 78 & 52 & \textbf{60.0} \\
\midrule
$\rho \to 0, N=2$ & 0 & - & - & - & - & - \\
$\rho=0.15, N=2$ & 62 & 49 & 53 & 74 & 48 & 57.2 \\
$\rho=0.25, N=2$ & 63 & 50 & 52 & 74 & 45 & 56.8 \\
\bottomrule
\end{tabular}
}
\end{table}

\begin{table}[htbp]
\centering
\caption{Comparison with 1-NFE Meanflow-based generative baselines.}
\label{tab:single_step_comparison}
\resizebox{\columnwidth}{!}{
\begin{tabular}{@{} l ccccc c @{}}
\toprule
\multirow{2}{*}{\textbf{Method}} & \multicolumn{5}{c}{\textbf{Task Success Rate (\%)}} & \multirow{2}{*}{\textbf{Average}} \\
\cmidrule(lr){2-6}
& \textbf{Pick} & \textbf{Diverse} & \textbf{Dual} & \textbf{Empty} & \textbf{Shoe} & \\
\midrule
Meanflow~\cite{geng2025mean} & 49 & 40 & 49 & 65 & 42 & 49 \\
IMF~\cite{geng2025improved} & 44 & 43 & 45 & 67 & 40 & 47.8 \\
Split-Meanflow~\cite{guo2025splitmeanflow} & 50 & 42 & 50 & 67 & 42 & 50.2 \\
\rowcolor{gray!15} \textbf{Ours} & \textbf{58} & \textbf{47} & \textbf{51} & \textbf{72} & \textbf{46} & \textbf{54.8} \\
\bottomrule
\end{tabular}
}
\end{table}

\textbf{Superiority over Single-Step Baselines.} To validate RCAF against existing 1-NFE acceleration techniques, we benchmark against the Meanflow family (Table~\ref{tab:single_step_comparison}). RCAF achieves an average success rate of $54.8\%$, outperforming the strongest single-step baseline, Split-Meanflow, by $4.6\%$ absolute improvement. These results support our hypothesis that directly learning an interval-averaged velocity can be limited when approximating nonlinear action trajectories. In contrast, RCAF estimates a recursive multi-step reference trajectory and applies residual correction to compensate for spatial deviation induced by long-span one-step approximation. This improves manipulation robustness while maintaining 1-NFE inference efficiency.

\begin{table}[htbp]
\centering
\caption{Comparison of frequency-domain constraint strategies.}
\label{tab:freq_constraint_comparison}
\resizebox{\columnwidth}{!}{
\begin{tabular}{@{} l ccccc c @{}}
\toprule
\multirow{2}{*}{\textbf{Method}} & \multicolumn{5}{c}{\textbf{Task Success Rate (\%)}} & \multirow{2}{*}{\textbf{Average}} \\
\cmidrule(lr){2-6}
& \textbf{Pick} & \textbf{Diverse} & \textbf{Dual} & \textbf{Empty} & \textbf{Shoe} & \\
\midrule
Wavelet Transform (db4) & 54.0 & 44.0 & 48.0 & 72.0 & 42.0 & 52.0 \\
Freqpolicy~\cite{su2025freqpolicy} & 56.0 & 46.0 & 51.0 & 76.0 & 40.0 & 53.8 \\
\midrule
\textbf{DTFC (w/o Curriculum)} & 60.0 & 49.0 & 51.0 & 74.0 & 45.0 & 55.8\\
\rowcolor{gray!15} \textbf{DTFC (Ours)} & \textbf{62.0} & \textbf{50.0} & \textbf{53.0} & \textbf{76.0} & \textbf{47.0} & \textbf{57.6} \\
\bottomrule
\end{tabular}
}
\end{table}

\textbf{Ablation on DTFC.} 
As shown in Table~\ref{tab:freq_constraint_comparison}, even without curriculum learning, the dual-timestep frequency consistency design outperforms both Freqpolicy~\cite{su2025freqpolicy} and wavelet-based frequency constraints. Adding the linear warmup curriculum further improves the average success rate to $57.6\%$, achieving the best performance among the compared frequency-domain strategies. These results suggest that a coarse-to-fine optimization schedule, which first learns the global action structure and then gradually strengthens high-frequency constraints, helps reduce early-stage gradient conflicts and improves high-frequency detail preservation.

\begin{table}[htbp]
\centering
\caption{Ablation of CFM margin strategies.}
\label{tab:cfm_ablation_flat}
\resizebox{\columnwidth}{!}{
\begin{tabular}{@{} l c c | c c c c c | c @{}}
\toprule
\textbf{Margin Type} & \textbf{Parameter} & $\lambda_{\text{cfm}}$ & \textbf{Pick} & \textbf{Diverse} & \textbf{Dual} & \textbf{Empty} & \textbf{Shoe} & \textbf{Avg.} \\
\midrule
Dynamic & $\alpha=1.0$ & 0.05 & 61.0 & 48.0 & 46.0 & 76.0 & 46.0 & 55.4 \\
Dynamic & $\alpha=1.0$ & 0.1 & 63.0 & 49.0 & 47.0 & 75.0 & 48.0 & 56.4 \\
\midrule
Static  & $m=0.4$ & 0.05 & 64.0 & 50.0 & 53.0 & \textbf{80.0} & 50.0 & 59.4 \\
\rowcolor{gray!15} \textbf{Static (Ours)} & \textbf{$m=0.4$} & \textbf{0.1} & \textbf{65.0} & \textbf{51.0} & \textbf{55.0} & 77.0 & \textbf{51.0} & \textbf{59.8} \\
\bottomrule
\end{tabular}
}
\end{table}
\textbf{Ablation on CFM Margin.} 
Table~\ref{tab:cfm_ablation_flat} compares a dynamic margin scaled by velocity discrepancies with a static margin. The dynamic margin underperforms in our experiments, likely because margin fluctuations introduce additional instability during early training. In contrast, the static margin with $m=0.4$ provides a stable repulsive constraint and achieves the best average success rate, suggesting that a fixed separation threshold is more reliable for separating entangled action flows.

\subsection{Real-World Robot Experiments}

To further evaluate physical deployment, we conduct real-world experiments on two robotic platforms, as shown on the right side of Figure~\ref{fig:benchmark_tasks}: the low-cost SO101 platform and an industrial UR7E collaborative robot. The SO101 platform uses a global camera and a wrist-mounted camera for visual feedback, while the UR7E platform is equipped with a RealSense D455 camera and a DM vision-tactile gripper, enabling evaluation across different robot embodiments, sensing configurations, and manipulation scales.

On SO101, we evaluate three manipulation tasks under low-data settings: \textit{Pick \& Place}, \textit{Stand Bottle}, and \textit{Drop Pen}. The \textit{Pick \& Place} task is further tested under both regular placement and position-generalization settings. On UR7E, we evaluate a stacking task and a position-generalization \textit{Pick \& Place} task to examine whether the learned policy can transfer to a larger industrial arm and handle object-position variations. Each setting is evaluated over 20 trials.

As shown in Table~\ref{tab:real_robot}, our method outperforms Diffusion Policy across both platforms. On SO101, we report a DP+RCAF variant and the complete DP+Ours framework to analyze the contribution of the proposed modules under low-data physical deployment. In the 30-demo position-generalization setting of \textit{Pick \& Place}, DP+Ours improves the success rate from 1/20 to 5/20. With 60 demonstrations, the gains become more pronounced, especially on challenging tasks such as \textit{Stand Bottle} and \textit{Drop Pen}. Compared with DP+RCAF, DP+Ours further improves performance on tasks requiring precise and stable manipulation, suggesting that DTFC and CFM provide additional benefits beyond spatial truncation-error correction.

On the UR7E platform, we evaluate our full method directly against Diffusion Policy. Our method improves the success rate from 4/20 to 7/20 on \textit{Stack Cylinder} and from 1/20 to 6/20 on the position-generalization \textit{Pick \& Place} task, indicating better robustness to unseen object placements. For completeness, we also evaluate $\pi_0$ in selected SO101 settings. While $\pi_0$ succeeds in one setting, some executions are aborted due to hardware safety triggers. Overall, these real-world results demonstrate that the proposed framework improves the reliability of accelerated visuomotor policies across different robot embodiments.

\begin{table}[htbp]
\centering
\caption{Real-world success rates on SO101 and UR7E. Each result is reported over 20 trials. ``Pos. Gen.'' denotes position generalization, ``Abort'' denotes executions stopped by hardware safety triggers, and ``N/A'' denotes settings not evaluated.}
\label{tab:real_robot}

\resizebox{\columnwidth}{!}{
\begin{tabular}{@{} l c c c c c @{}}
\toprule
\multicolumn{6}{c}{\textbf{SO101 Platform}} \\
\midrule
\textbf{Task} & \textbf{Demos} 
& \textbf{Diffusion Policy} & \textbf{DP+RCAF} & \textbf{DP+Ours} & \textbf{$\pi_0$} \\
\midrule
Pick \& Place (Pos. Gen.) & 30 & 1/20 & 3/20 & \textbf{5/20} & Abort \\
Pick \& Place (Pos. Gen.) & 60 & 4/20 & 7/20 & \textbf{10/20} & 6/20 \\
Pick \& Place (Regular) & 60 & 12/20 & \textbf{20/20} & \textbf{20/20} & Abort \\
Stand Bottle (Regular) & 60 & 2/20 & 14/20 & \textbf{18/20} & N/A \\
Drop Pen (Regular) & 60 & 8/20 & 8/20 & \textbf{12/20} & N/A \\
\bottomrule
\end{tabular}
}

\vspace{1mm}

\resizebox{\columnwidth}{!}{
\begin{tabular}{@{} l c c c @{}}
\toprule
\multicolumn{4}{c}{\textbf{UR7E Platform}} \\
\midrule
\textbf{Task} & \textbf{Demos} 
& \textbf{Diffusion Policy} & \textbf{Ours} \\
\midrule
Stack Cylinder & 80 & 4/20 & \textbf{7/20} \\
Pick \& Place (Pos. Gen.) & 80 & 1/20 & \textbf{6/20} \\
\bottomrule
\end{tabular}
}

\vspace{-2mm}
\end{table}

\section{Conclusion}
\label{sec:conclusion}

This paper presents a high-fidelity one-step generative visuomotor policy framework for efficient robotic manipulation. The framework integrates recursive correction, frequency consistency, and contrastive flow matching to address three key challenges in one-step action generation: spatial truncation, frequency distortion, and multimodal flow entanglement. Experiments across 33 simulation tasks and real-world robot platforms show that the proposed framework achieves competitive or superior performance compared with strong multi-step generative policy baselines while requiring only one function evaluation. These results suggest that jointly improving recursive spatial correction, frequency-domain consistency, and multimodal flow separation is important for high-fidelity single-step visuomotor control.

\textbf{Limitations and Future Work.}
Although our framework shows strong empirical performance, it introduces module-specific hyperparameters, such as recursive steps, frequency weights, and contrastive margins, which may require tuning when transferred to new robot embodiments or task distributions. Future work will explore scaling this single-step acceleration paradigm to larger multimodal robot foundation models and integrating tactile feedback for more dynamic contact-rich manipulation.

\appendix

\subsection{Experimental Setup}
\label{sec:appendix_impl}

To ensure the strict reproducibility of our framework, Table~\ref{tab:hyperparams} details the exhaustive hyperparameter configurations, network architectures, and hardware specifications utilized across all benchmarks. 

\begin{table}[htbp]
    \centering
    \caption{Detailed hyperparameters and implementation configurations across simulation benchmarks.}
    \label{tab:hyperparams}
    \resizebox{\columnwidth}{!}{
    \begin{tabular}{@{} lcc @{}}
        \toprule
        \textbf{Parameter} & \textbf{RoboTwin} & \textbf{Adroit \& DexArt} \\
        \midrule
        \multicolumn{3}{c}{\textit{Hardware \& Software Specifications}} \\
        \midrule
        Compute Hardware & \multicolumn{2}{c}{NVIDIA RTX 4090} \\
        Deep Learning Framework & \multicolumn{2}{c}{PyTorch 2.4.1 (CUDA 11.8)} \\
        \midrule
        \multicolumn{3}{c}{\textit{Network Architecture}} \\
        \midrule
        Visual Encoder & \multicolumn{2}{c}{R3M (ResNet-18) / DP3} \\
        Policy Backbone & \multicolumn{2}{c}{DiTX (12 layers, 8 heads, 768 dim)} \\
        Diffusion Timestep Embed Dim & \multicolumn{2}{c}{128} \\
        \midrule
        \multicolumn{3}{c}{\textit{General Training Configurations}} \\
        \midrule
        Optimizer & \multicolumn{2}{c}{AdamW (Base LR: $5 \times 10^{-5}$)} \\
        Batch Size & 64 & 192 \\
        Training Epochs & 1010 (2D) / 2010 (3D) & 5010 \\
        Observation Horizon ($T_{\text{obs}}$) & \multicolumn{2}{c}{2} \\
        Prediction Horizon ($T_{\text{pred}}$) & 16 & 4 \\
        Action Execution Steps ($T_{\text{act}}$) & 16 & 3 \\
        \midrule
        \multicolumn{3}{c}{\textit{Framework-Specific Settings}} \\
        \midrule
        Flow / Consistency Batch Ratio & \multicolumn{2}{c}{0.75 / 0.25} \\
        Recursive Steps ($N$) & \multicolumn{2}{c}{2} \\
        Anchor Step Ratio ($\rho$) & \multicolumn{2}{c}{0.2} \\
        Residual Clamp Magnitude ($\gamma$) & \multicolumn{2}{c}{5.0} \\
        Freq. Consistency Weight ($\lambda_{\text{freq}}$) & \multicolumn{2}{c}{0.2} \\
        Focal Weight Alpha ($\alpha$) & \multicolumn{2}{c}{2.0} \\
        Gaussian Mask Bandwidth ($\sigma$) & \multicolumn{2}{c}{0.3} \\
        Curriculum Warmup Epochs ($E_{\text{warmup}}$) & \multicolumn{2}{c}{200} \\
        Contrastive Margin / Weight & \multicolumn{2}{c}{$m=0.4$ / $\lambda_{\text{CFM}}=0.1$} \\
        EMA Decay Rate ($\beta$) & \multicolumn{2}{c}{0.9999} \\
        \bottomrule
    \end{tabular}
    }
\end{table}

\subsection{Training Dynamics and Convergence Analysis}
\label{sec:appendix_training_curves}
\begin{figure*}[htbp]
    \centering
    \includegraphics[width=0.75\textwidth]{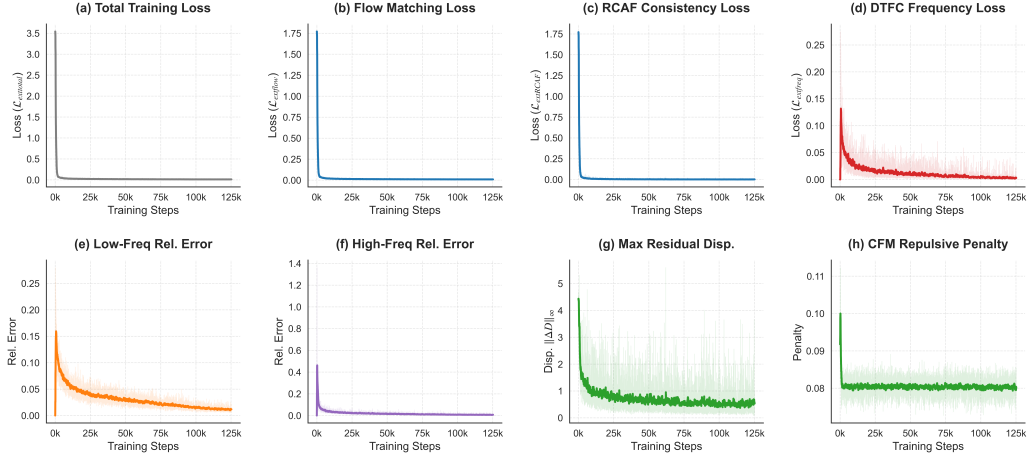}
    \caption{Training dynamics and convergence analysis logged over 120k steps. The subplots systematically evaluate (a) the total loss, (b-d) the convergence of the three core sub-objectives, (e-f) the synchronous recovery of cross-spectral relative errors, and (g-h) the bounded stabilization of the internal structural constraints (the RCAF spatial clamp $\xi$ and the CFM margin penalty).}
    \label{fig:training_curves}
\end{figure*}
Figure~\ref{fig:training_curves} visualizes the training dynamics, validating the optimization stability of our framework without numerical oscillations or gradient explosion. 

\textbf{Core Objectives Convergence:} 
(a) Total Loss converges smoothly after an initial rapid descent. 
(b) Flow Matching Loss drops rapidly, effectively establishing the base action manifold topology. 
(c) RCAF Consistency Loss decays synchronously with the flow loss, confirming the stable distillation of high-order residual corrections from the EMA teacher. 
(d) DTFC Frequency Loss steadily declines under the curriculum warmup, bridging the spectral gap across timesteps.

Spectral Fidelity \& Internal Constraints: Relative errors in both (e) Low-Frequency and (f) High-Frequency bands decay synchronously to $\sim 1.0$. This verifies that our dynamic focal weighting successfully prevents low-pass degradation, capturing both global macro-motions and transient high-frequency details. 

Critically, (g) the Maximum Residual Displacement ($\xi$) drops dramatically from an initial peak of $\sim 4.7$ to near zero. This signifies that the single-step student naturally internalizes the manifold curvature as training progresses, aligning its linear predictions with the true multi-step trajectories. Finally, (h) the CFM Repulsive Penalty stabilizes strictly around the defined margin, providing a continuous, non-destructive decoupling force for intertwined action flows.

\subsection{Phenomenon-Analysis}
\begin{figure*}[t]
    \centering
    \includegraphics[width=0.8\textwidth]{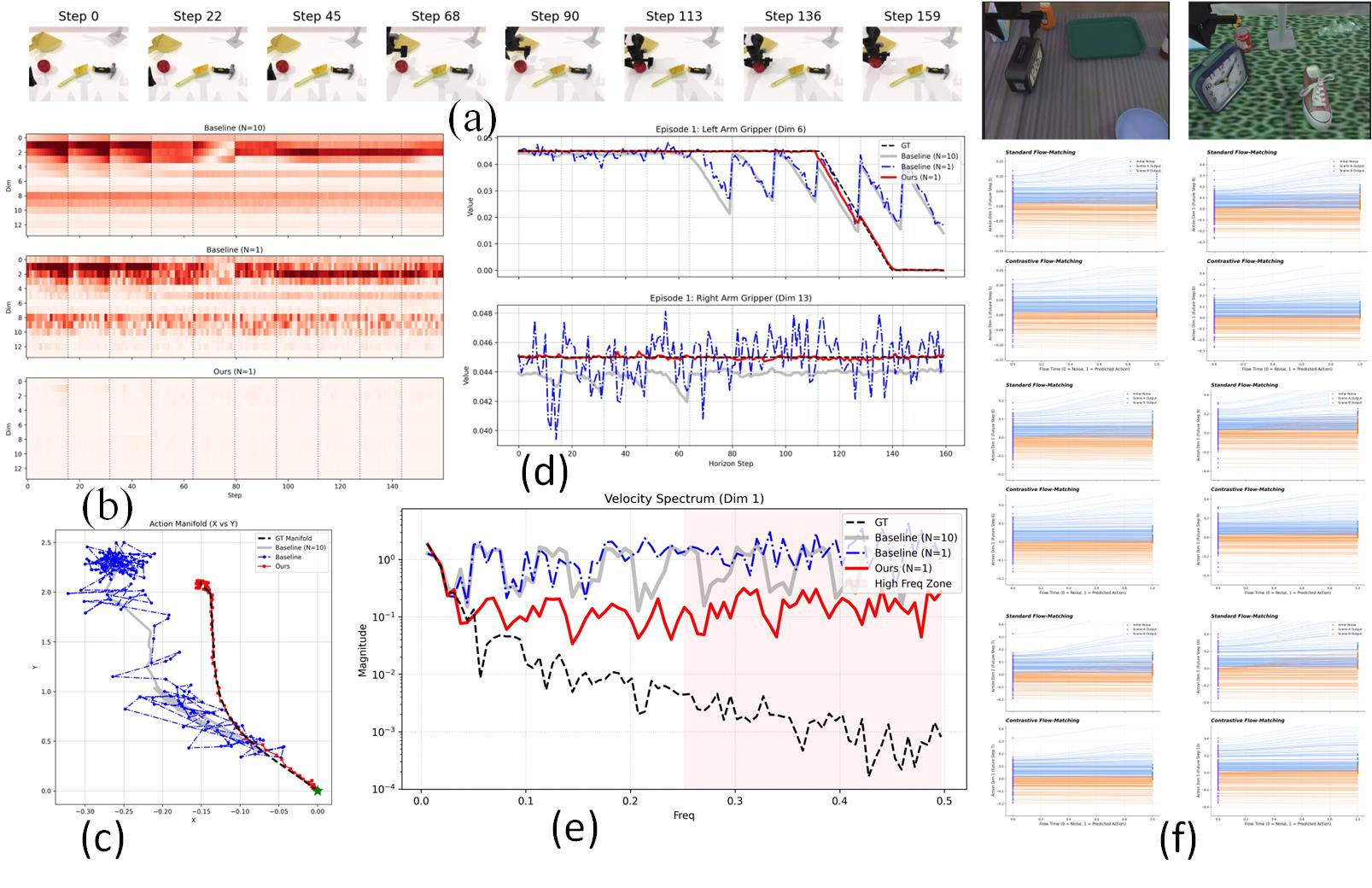}
    \caption{
        \textbf{Comprehensive Qualitative Analysis and CFM Empirical Verification.} 
        \textbf{(a-e) Action Generation Stability:} (a) RGB observation sequence; (b) Absolute action error heatmaps; (c) 2D phase space action manifold; (d) Time-domain trajectories of grippers; (e) Frequency-domain velocity spectrum demonstrating high-frequency noise suppression. 
        \textbf{(f) Empirical Verification of CFM:} Trajectories of 100 identical Gaussian noise samples evolving to predicted actions for two distinct scenes. Standard flow-matching (top rows) suffers from mode-averaging entanglement, whereas our CFM (bottom rows) establishes a latent repulsive field, encouraging more separated, decoupled, and scene-specific action manifolds.
    }
    \label{fig:unified_qualitative}
\end{figure*}

To explicitly evaluate the stability and discriminative capacity of our approach in the challenging single-step regime ($N=1$), we visualize the comprehensive action generation process in Figure~\ref{fig:unified_qualitative}.

\textbf{Action Stability and Spectral Fidelity.} As demonstrated by the absolute error heatmaps (Fig.~\ref{fig:unified_qualitative}b), the conventional 1-NFE baseline suffers from larger systematic errors. Physically, this degradation manifests as severe geometric jittering—evident in the less structured 2D action manifold (Fig.~\ref{fig:unified_qualitative}c) and the temporal chattering of the grippers (Fig.~\ref{fig:unified_qualitative}d). Conversely, our generated trajectory tightly adheres to the expert ground-truth (GT) manifold, executing sharp state transitions without physical oscillation. Frequency-domain analysis (Fig.~\ref{fig:unified_qualitative}d,e) provides one explanation for this gap: our mechanism successfully suppresses anomalous high-frequency noise by over an order of magnitude compared to the baseline.

\textbf{Empirical Verification of CFM Decoupling.} To visualize how Contrastive Flow Matching (CFM) mitigates the mode-averaging bottleneck, Figure~\ref{fig:unified_qualitative}f traces the continuous integration of 100 identical Gaussian noise samples evolving toward the action space for two visually distinct scenes. Without explicit contrastive constraints (Fig.~\ref{fig:unified_qualitative}f, top rows), standard flow-matching struggles with overlapping state-action spaces. Trajectories from different scenes entangle and collapse toward an ambiguous mean center, inducing high ODE curvature and amplifying single-step truncation errors. In stark contrast, our CFM-equipped policy (Fig.~\ref{fig:unified_qualitative}f, bottom rows) constructs a latent repulsive field. This cleanly decouples the previously interwoven flows into independent manifolds. The evolutionary paths remain straight and deterministically map identical initial noises to distinct, scene-specific action distributions, thereby guaranteeing execution precision.

\subsection{Theoretical Analysis of Recursive Consistent Action Flow}
\label{sec:theoretical_analysis}

To deeply understand the optimization dynamics of RCAF, we provide a theoretical analysis from a control theory perspective.

\textbf{Implicit High-Gain Feedback.} Let $\Delta t$ be the integration span, $dt_{\text{anc}} = \rho \cdot \Delta t$ the anchor duration, $v_{\text{long}}$ the teacher's global velocity, $D_{\text{rec}}$ the $N$-step accumulated displacement, and $v_{\text{true}}$ the physical ground-truth velocity. The target velocity for the student policy is defined by the residual correction $\xi = v_{\text{est}} - v_{\text{true}}$:
\begin{equation}
v_{\text{target}} = v_{\text{long}} - \underbrace{\left( \frac{v_{\text{long}} \Delta t - D_{\text{rec}}}{dt_{\text{anc}}} \right)}_{v_{\text{est}}} + v_{\text{true}}
\end{equation}
Defining the scaling constant $k = \Delta t / dt_{\text{anc}} = 1/\rho$, this expands to:
\begin{equation}
v_{\text{target}} = (1 - k) v_{\text{long}} + k \frac{D_{\text{rec}}}{\Delta t} + v_{\text{true}}
\end{equation}
This reveals RCAF intrinsically acts as a High-Gain Proportional Controller. For our default $\rho = 0.2$ ($k = 5$), the coefficient for $v_{\text{long}}$ is $-4$, meaning prediction deviations are inversely amplified by a factor of $4$. This aggressive negative feedback strongly penalizes deviations from the curved manifold, forcing the policy away from erroneous linear shortcuts.

\textbf{Fixed-Point Convergence.} Despite aggressive amplification, the system suggests a stable fixed point under the assumed formulation where the student matches the target and the EMA teacher ($v_{\text{target}} = v_{\text{long}}$), yielding $\xi \to 0$. Substituting this into the expansion and recalling $1/k = dt_{\text{anc}} / \Delta t$, we arrive at the analytical solution:
\begin{equation}
v_{\text{long}} = \frac{v_{\text{true}} \cdot dt_{\text{anc}} + D_{\text{rec}}}{\Delta t}
\end{equation}
This mathematical fixed point is exactly the Global Average Secant Velocity across the action manifold, ensuring the model aligns with the true multi-step integration path.

\textbf{Gradient Magnification.} Let $v_{\text{opt}}$ denote the ideal secant velocity derived above. We can algebraically reformulate the target as $v_{\text{target}} = v_{\text{long}} - k(v_{\text{long}} - v_{\text{opt}})$. Minimizing the standard MSE objective $\mathcal{L} = \frac{1}{2} \| v_{\text{student}} - \text{sg}(v_{\text{target}}) \|^2$ yields the gradient:
\begin{equation}
\nabla_{v_{\text{student}}} \mathcal{L} = (v_{\text{student}} - v_{\text{long}}) + k(v_{\text{long}} - v_{\text{opt}})
\end{equation}
Since the EMA teacher's local predictions are highly correlated with the student ($v_{\text{student}} \approx v_{\text{long}}$), the effective gradient driving the optimization simplifies to:
\begin{equation}
\nabla_{v_{\text{student}}} \mathcal{L} \approx k \cdot (v_{\text{long}} - v_{\text{opt}})
\end{equation}


\bibliographystyle{ieeetr} 
\bibliography{ref} 


\end{document}